\documentclass[10pt,twocolumn,letterpaper]{article}

\usepackage{cvpr}
\usepackage{times}
\usepackage{epsfig}
\usepackage{graphicx}
\usepackage{amsmath}
\usepackage{amssymb}

\usepackage[colorlinks]{hyperref}
\usepackage{color}
\usepackage{multirow}
\usepackage{nicefrac}       
\usepackage{microtype}      
\usepackage{booktabs}       

\newcommand{\norm}[1]{\left\|#1\right\|} 
\newcommand{\inner}[2]{#1\cdot #2} 
\newcommand*\rot{\rotatebox{90}}

\cvprfinalcopy 

\def\cvprPaperID{197} 

\ifcvprfinal\fi
\begin{document}

\title{Taking A Closer Look at Domain Shift: \\ Category-level Adversaries for Semantics Consistent Domain Adaptation}

\author{Yawei Luo$^{1,2}$,\hspace{2mm} Liang Zheng$^{5}$,\hspace{2mm} Tao Guan$^{1,6}$,\hspace{2mm} Junqing Yu$^{1,4}$ \thanks{Corresponding author (yjqing@hust.edu.cn). 
\newline \hspace*{0.16in} This work was done when Yawei Luo (royalvane@hust.edu.cn) was a visiting student at University of Technology Sydney. Part of this work was done when Yi Yang (yee.i.yang@gmail.com) was visiting Baidu Research during his Professional Experience Program. The code is publicly available at \url{https://github.com/RoyalVane/CLAN}. 
},\hspace{2mm} Yi Yang$^{2,3}$ \vspace{0.5cm} \\ 
$^1$School of Computer Science \& Technology, Huazhong University of Science \& Technology\\
$^2$CAI, University of Technology Sydney ~ $^3$Baidu Research\\ 
$^4$Center of Network and Computation, Huazhong University of Science \& Technology \\
$^5$Research School of Computer Science, Australian National University ~ $^6$Farsee2 Tech. Co.
}

\maketitle
\vspace{-0.2cm}
\begin{abstract}
   We consider the problem of unsupervised domain adaptation in semantic segmentation. A key in this campaign consists in reducing the domain shift, i.e., enforcing the data distributions of the two domains to be similar. One of the common strategies is to align the marginal distribution in the feature space through adversarial learning. However, this global alignment strategy does not consider the category-level joint distribution. A possible consequence of such global movement is that some categories which are originally well aligned between the source and target may be incorrectly mapped, thus leading to worse segmentation results in target domain. To address this problem, we introduce a category-level adversarial network, aiming to enforce local semantic consistency during the trend of global alignment. Our idea is to take a close look at the category-level joint distribution and align each class with an adaptive adversarial loss. Specifically, we reduce the weight of the adversarial loss for category-level aligned features while increasing the adversarial force for those poorly aligned. In this process, we decide how well a feature is category-level aligned between source and target by a co-training approach. In two domain adaptation tasks, i.e., GTA5 $\rightarrow$ Cityscapes and SYNTHIA $\rightarrow$ Cityscapes, we validate that the proposed method matches the state of the art in segmentation accuracy. 
\end{abstract}
\vspace{-0.6cm}

\begin{figure}[t]
\centering
\includegraphics[width=0.99\linewidth]{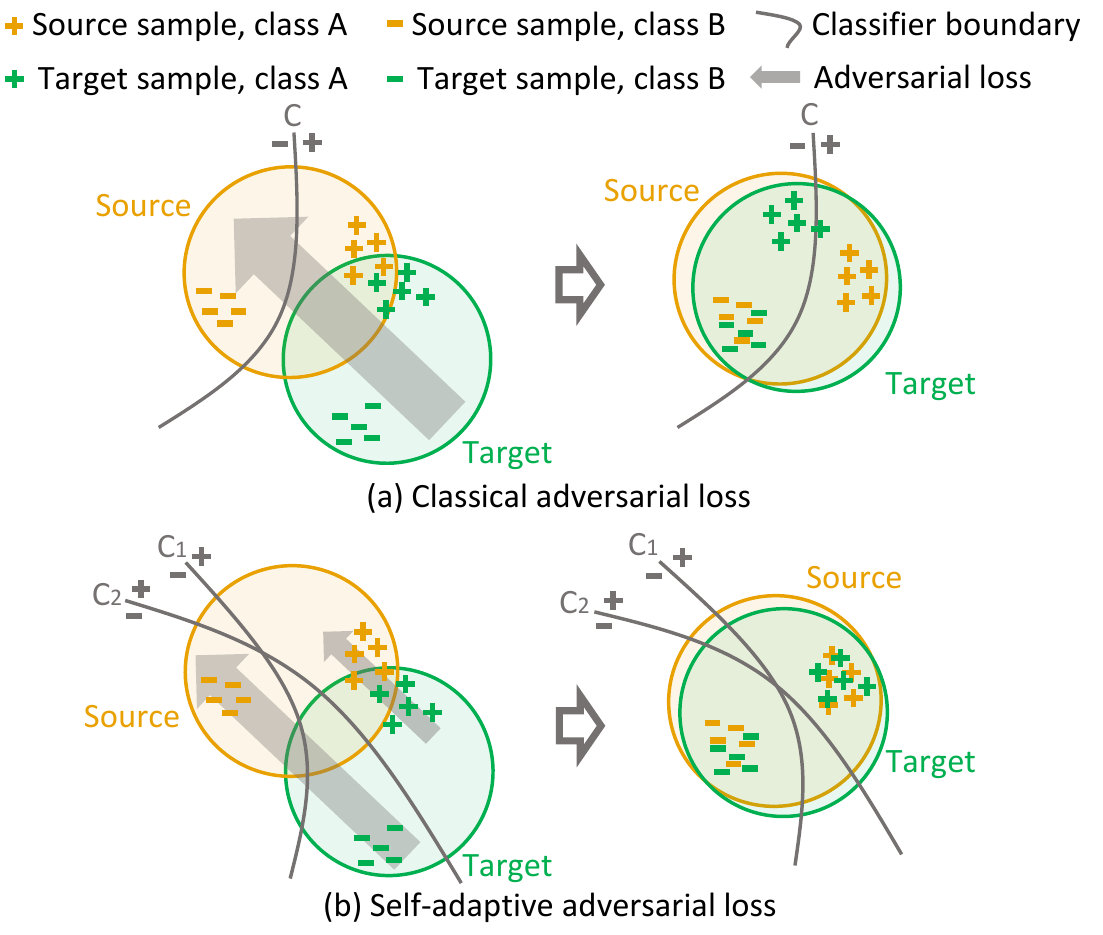}
\caption{(Best viewed in color.) Illustration of traditional and the proposed adversarial learning. The size of the solid gray arrow represents the weight of the adversarial loss. (a) Traditional adversarial learning ignores the semantic consistency when pursuing the marginal distribution alignment. As a result, the global movement might cause the well-aligned features (class A) to be mapped onto different joint distributions (negative transfer). (b) The proposed self-adaptive adversarial learning reweights the adversarial loss for each feature by a local alignment score. Our method reduces the influence of the adversaries when discovers a high semantic alignment score on a feature, and vice versa. As is shown, the proposed strategy encourages a category-level joint distribution alignment for both class A and class B.}
\label{fig:Brief} 
\vspace{-2mm}
\end{figure}
\section{Introduction}
Semantic segmentation aims to assign each pixel of a photograph to a semantic class label. Currently, the achievement is at the price of large amount of dense pixel-level annotations obtained by expensive human labor ~\cite{chen2018deeplab,long2015fcn,luo2018macro}. An alternative would be resorting to simulated data, such as computer generated scenes~\cite{richter2016gta5,ros2016synthia}, so that unlimited amount of labels are made available. However, models trained with the simulated images do not generalize well to realistic domains. The reason lies in the different data distributions of the two domains, typically known as domain shift~\cite{shimodaira2000improving}. To address this issue, domain adaptation approaches~\cite{saito2017maximum,tsai2018OutputSpace,hoffman2017cycada,zhong2019invariance,kang2018deep,kang2019contrastive,he2019filter,zhu2019simreal} are proposed to bridge the gap between the source and target domains. 
A majority of recent methods~\cite{long2015MMD,sun2016deep,tzeng2017adversarial,tzeng2015simultaneous} aim to align the feature distributions of different domains. Works along this line are based on the theoretical insights in~\cite{ben2010theory} that minimizing the divergence between domains lowers the upper bound of error on the target domain. Among this cohort of domain adaptation methods, a common and pivotal step is minimizing some distance metric between the source and target feature distributions \cite{long2015MMD,sun2016deep}. Another popular choice, which borrows the idea from adversarial
learning~\cite{goodfellow2014gan}, is to minimize the accuracy of domain prediction. Through a minimax game between two adversarial networks, the generator is trained to produce features that confuse the discriminator while the latter is required to correctly classify which domain the features are generated from. 

Although the works along the path of adversarial learning have led to impressive results~\cite{solomon2015WGAN,hoffman2016fcns,liu2016coupled,kim2017relations,tzeng2017adversarial,sankaranarayanan2017unsupervised}, they suffer from a major limitation: when the generator network can perfectly fool the discriminator, it merely aligns the global marginal distribution of the features in the two domains ( \emph{i.e.}, $P(F_s) \approx P(F_t)$, where $F_s$ and $F_t$ denote the features of source and target domain in latent space) while ignores the local joint distribution shift, which is closely related to the semantic consistency of each category (\emph{i.e.}, $P(F_s, Y_s) \neq P(F_t, Y_t)$, where $Y_s$ and $Y_t$ denote the categories of the features).  
As a result, the \emph{de facto} use of the adversarial loss may cause those target domain features, which are already well aligned to their semantic counterpart in source domain, to be mapped to an incorrect semantic category (negative transfer). This side effect becomes more severe when utilize a larger weight on the adversarial loss. 

To address the limitation of the global adversarial learning, we propose a category-level adversarial network (CLAN),  prioritizing category-level alignment which will naturally lead to global distribution alignment. The cartoon comparison of traditional adversarial learning and the proposed one is shown in Fig.~\ref{fig:Brief}. The key idea of CLAN is two-fold. First, we identify those classes whose features are already well aligned between the source and target domains, and protect this category-level alignment from the side effect of adversarial learning. Second, we identify the classes whose features are distributed differently between the two domains and increase the weight of the adversarial loss during training. In this process, we utilize co-training \cite{zhou2005cotrain}, which enables high-confidence predictions with two diverse classifiers, to predict how well each feature is semantically aligned between the source and target domains. Specifically, if the two classifiers give consistent predictions, it indicates that the feature is predictive and achieves good semantic alignment. In such case, we reduce the influence of the adversarial loss in order to encourage the network to generate invariant features that can keep semantic consistency between domains. On the contrary, if the predictions disagree with each other, which indicates that the target feature is far from being correctly mapped, we increase the weight of the adversarial loss on that feature so as to accelerate the alignment. Note that 1) Our adversarial learning scheme acts directly on the output space. By regarding the output predictions as features, the proposed method jointly promotes the optimization for both classifier and extractor; 2) Our method does not guarantee rigorous joint distribution alignment between domains. Yet, compared with marginal distribution alignment, our method can \emph{map the target features closer (or no negative transfer at worst)} to the source features of the same categories. The main contributions are summarized below.

\begin{itemize}
\item By proposing to adaptively weight the adversarial loss for different features, we emphasize the importance of category-level feature alignment in reducing domain shift. 
\item Our results are on par with the state-of-the-art UDA methods on two transfer learning tasks, \emph{i.e.}, GTA5~\cite{richter2016gta5} $\rightarrow$ Cityscapes~\cite{cordts2016cityscapes} and SYNTHIA~\cite{ros2016synthia} $\rightarrow$ Cityscapes.
\end{itemize}

\section{Related Works}
This section will focus on adversarial learning and co-training techniques for unsupervised domain adaptation, which form the two main motivations of our method.

\textbf{Adversarial learning.}
Ben-David \emph{et al.}~\cite{ben2010theory} had proven that the adaptation loss is bounded by three terms, \emph{e.g.}, the expect loss on source domain, the domain divergence, and the shared error of the ideal joint hypothesis on the source and target domain. Because the first term corresponds to the well-studied supervised learning problems and the third term is considered sufficiently low to achieve an accurate adaptation, the majority of recent works lay emphasis on the second term. Adversarial adaptation methods are good examples of this type of approaches and can be investigated on different levels. Some methods focus on the distribution shift in the latent feature space \cite{solomon2015WGAN,hoffman2016fcns,liu2016coupled,kim2017relations,tzeng2017adversarial,sankaranarayanan2017unsupervised}. In an example, Hoffman \emph{et al.}~\cite{hoffman2016fcns} appended category statistic constraints to the adversarial model, aiming to improve semantic consistency in target domain. Other methods address the adaption problem on the pixel level \cite{li2018Grad-GAN,bousmalis2017pixelDA}, which relate to the style transfer approaches~\cite{zhu2017cycle,choi2017stargan} to make images indistinguishable across domains. A joint consideration of pixel and feature level domain adaptation is studied in~\cite{hoffman2017cycada}. Besides alignment in the bottom feature layers, Tsai \emph{et al.}~\cite{tsai2018OutputSpace} found that aligning directly the output space is more effective in semantic segmentation. Domain adaptation in the output space enables the joint optimization for both prediction and representation, so our method utilizes this advantage. 

\textbf{Co-training.} 
Co-training~\cite{zhou2005cotrain} belongs to multi-view learning in which learners are trained alternately on two distinct views with confident labels from the unlabeled data. In UDA, this line of methods~\cite{zhang2017FCTN, chen2011co, saito2017asymmetric,luo2008transfer} are able to assign pseudo labels to unlabeled samples in the target domain, which enables direct measurement and minimization the classification loss on target domain. 
In general, co-training enforces the two classifiers to be diverse in the learned parameters, which can be achieved via dropout~\cite{saito2017dropout}, consensus regularization~\cite{saito2017maximum} or parameter diverse~\cite{zhang2017FCTN}, \emph{etc}. Similar to co-training, tri-training keeps the two classifiers producing pseudo labels and uses these pseudo labels to train an extra classifier~\cite{saito2017asymmetric,zhang2017FCTN}. Apart from assigning pseudo labels to unlabeled data, Saiko \emph{et al.}~\cite{saito2017dropout,saito2017maximum} maximized the consensus of two classifiers for domain adaptation. 

Our work does not follow the strategy of global feature alignment~\cite{tsai2018OutputSpace, hoffman2016fcns, solomon2015WGAN} or classifiers consensus maximization~\cite{saito2017dropout,saito2017maximum}. Instead, category-level feature alignment is enforced through co-training.  
To our knowledge, we are making an early attempt to adaptively weight the adversarial loss for features in segmentation task according to the local alignment situation. 

\begin{figure*}[ht]
\centering
\includegraphics[width=0.95\linewidth]{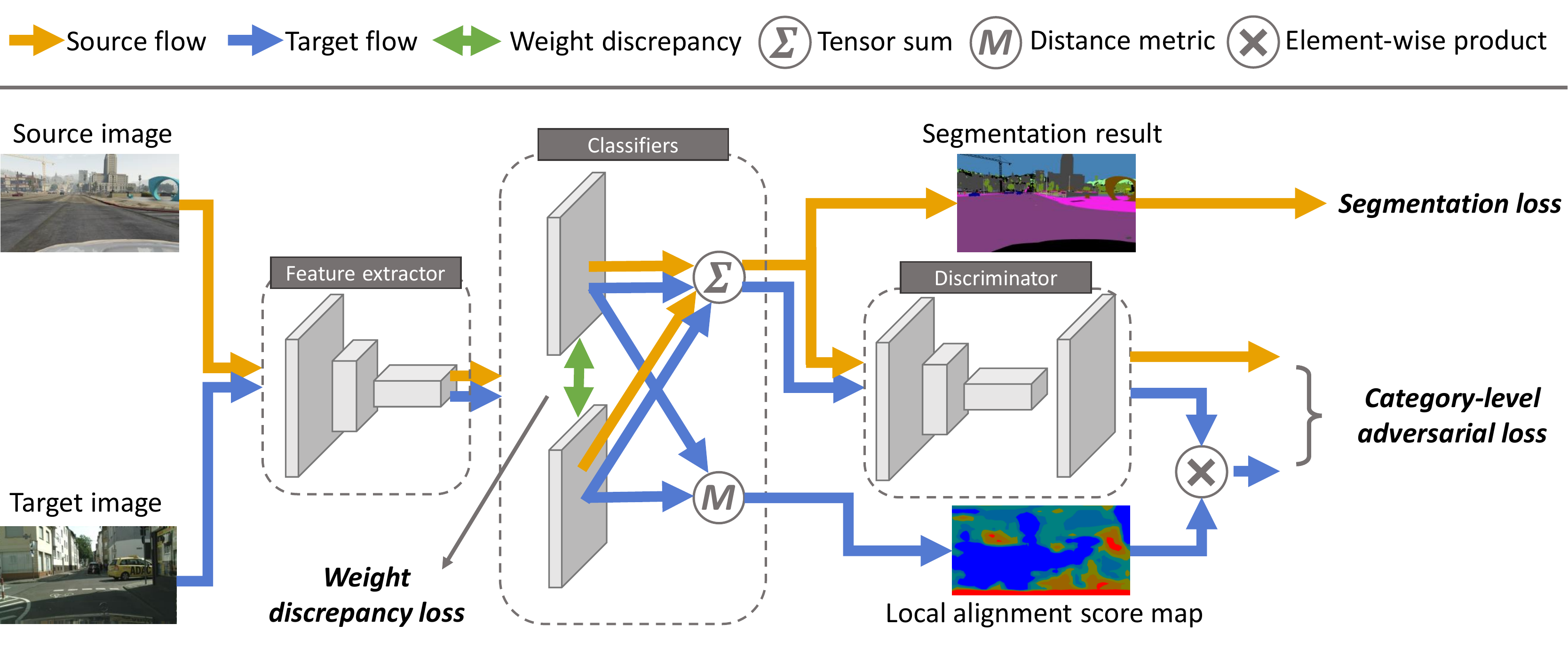}
\caption{Overview of the proposed category-level adversarial network. It consists of a feature extractor $E$, two classifiers $C_1$ and $C_2$, and a discriminator $D$. $C_1$ and $C_2$ are fed with the deep feature map extracted from $E$ and predict semantic labels for each pixel from diverse views. In source flow, the sum of the two prediction maps is used to calculate a segmentation loss as well as an adversarial loss from $D$. In target flow, the sum of the two prediction maps is forwarded to $D$ to produce a \emph{raw} adversarial loss map. Additionally, we adopt the discrepancy of the two prediction maps to produce a local alignment score map. This map evaluates the category-level alignment degree of each feature and is used to adaptively weight the raw adversarial loss map.}
\label{fig:main}
\vspace{-1mm}
\end{figure*}

\section{Method}

\subsection{Problem Settings} \label{Problem setting}
We focus on the problem of unsupervised domain adaptation (UDA) in semantic segmentation, where we have access to the source data $X_S$ with pixel-level labels $Y_S$, and the target data $X_T$ without labels. The goal is to learn a model $G$ that can correctly predict the pixel-level labels for the target data $X_T$. Traditional adversaries-based networks (TAN) consider two aspects for domain adaptation. First, these methods train a model $G$ that distills knowledge from labeled data in order to minimize the segmentation loss in the source domain, formalized as a fully supervised problem:

\begin{equation}
\mathcal{L}_{seg} (G) = E[ \ell (G(X_S), Y_S) ] \; ,
\label{eq-basic}
\end{equation}
where $E[ \cdot ]$ denotes statistical expectation and $\ell(\cdot, \cdot)$ is an appropriate loss function, such as multi-class cross entropy.

Second, adversaries-based UDA methods also train $G$ to learn domain-invariant features by confusing a domain discriminator $D$ which is able to distinguish between samples of the source and target domains. This property is achieved by minimaxing an adversarial loss:

\begin{equation}
\begin{aligned}
\mathcal{L}_{adv} (G, D) = & -E[ \log (D(G(X_S)))] \\
& -E[ \log (1-D(G(X_T))) ] \; .
\label{eq-basic-adversarial}
\end{aligned}
\end{equation}

However, as mentioned above, there is a major limitation for traditional adversarial learning methods: even under perfect alignment in marginal distribution, there might be the negative transfer that causes the samples from different domains but of the same class label to be mapped farther away in the feature space. In some cases, some classes are already aligned between domains, but the adversarial loss might deconstruct the existing local alignment when pursuing the global marginal distribution alignment. In this paper, we call this phenomenon ``lack of  semantic consistency'', which is a critical cause of performance degradation.

\subsection{Network Architecture} \label{Network architecture}
Our network architecture is illustrated in Fig.~\ref{fig:main}. It is composed of a generator $G$ and a discriminator $D$. $G$ can be any FCN-based segmentation network~\cite{simonyan2014vgg, long2015fcn, chen2018deeplab} and $D$ is a CNN-based binary classifier with a fully-convolutional output~\cite{goodfellow2014gan}. 
As suggested in the standard co-training algorithm~\cite{zhou2005cotrain}, generator $G$ is divided into feature extractor $E$ and two classifiers $C_1$ and $C_2$. $E$ extracts features from input images; $C_1$ and $C_2$ classify features generated from $E$ into one of the pre-defined semantic classes, such as car, tree and road. Following the co-training practice, we enforce the weights of $C_1$ and $C_2$ to be diverse through a cosine distance loss. This will provide us with the distinct views / classifiers to make semantic predictions for each feature. The final prediction map $p$ is obtained by summing up the two diverse prediction tensors $p^{(1)}$ and $p^{(2)}$ and we call $p$ an \emph{ensemble prediction}.

Given a source domain image $x_s \in X_S$, feature extractor $E$ outputs a feature map, which is input to classifiers $C_1$ and $C_2$ to yield the pixel-level ensemble prediction $p$. On the one hand, $p$ is used to calculate a segmentation loss under the supervision of the ground-truth label $y_s \in Y_S$. On the other hand, $p$ is input to $D$ to generate an adversarial loss. 

Given a target domain image $x_t \in X_T$, we also forward it to $G$ and obtain an ensemble prediction $p$. Different from the source data flow, we additionally generate a discrepancy map out of $p^{(1)}$ and $p^{(2)}$, denoted as $\mathcal{M}(p^{(1)}, p^{(2)})$, where $\mathcal{M}( \cdot, \cdot )$ denotes some proper distance metric to measure the element-wise discrepancy between $p^{(1)}$ and $p^{(2)}$. When using the cosine distance as an example, $\mathcal{M}(p^{(1)}, p^{(2)})$ forms a $1 \times H \times W$ shaped tensor with the $(i_{th} \in H, j_{th} \in W)$ element equaling to $(1 - \cos(p^{(1)}_{i,j}, p^{(2)}_{i,j}))$. Once $D$ produces an adversarial loss map $\mathcal{L}_{adv}$, an element-wise multiplication is performed between $\mathcal{L}_{adv}$ and $\mathcal{M}(p^{(1)}, p^{(2)})$. As a result, the final adaptive adversarial loss on a target sample takes the form as $\sum_{i=1}^H \sum_{j=1}^W (1 - \cos(p^{(1)}_{i,j}, p^{(2)}_{i,j})) \times \mathcal{L}_{adv_{i,j}}$, where $\{i, j\}$ traverses over all the pixels on the map. In this manner, each pixel on the segmentation map is differently weighted \emph{w.r.t} the adversarial loss.

\subsection{Training Objective} \label{Training objective}
The proposed network is featured by three loss functions, \emph{i.e.}, the \emph{segmentation loss}, the \emph{weight discrepancy loss} and the \emph{self-adaptive adversarial loss}. Given an image $x \in X_S$ of shape $3\times H\times W$  and a label map $y \in Y_S$ of shape $C\times H\times W$ where $C$ is the number of semantic classes, the segmentation loss (multi-class cross-entropy loss) can be concretized from Eq. \ref{eq-basic} as
\begin{equation}
	\mathcal{L}_{seg}(G) = \sum_{i=1}^{H\times W}\sum_{c=1}^{C} - y_{ic}\log{p_{ic}} \; ,
\end{equation}
where ${p}_{ic}$ denotes the predicted probability of class $c$ on pixel $i$. $y_{ic}$ denotes the ground truth probability of class $c$ on the pixel $i$. If pixel $i$ belongs to class $c$, $y_{ic}=1,$ otherwise $y_{ic}=0$.

For the second loss, as suggested in the standard co-training algorithm~\cite{zhou2005cotrain}, the two classifiers $C_1$ and $C_2$ should have possibly diverse parameters in order to provide two different views on a feature. Otherwise, the training degenerates to self-training. Specifically, we enforce divergence of the weights of the convolutional layers of the two classifiers by \emph{minimizing} their cosine similarity. Therefore, we have the following weight discrepancy loss:
\begin{equation}
\mathcal{L}_{weight} (G) = \frac{\inner{\vec{w_1}}{\vec{w_2}}}{\norm{\vec{w_1}}\norm{\vec{w_2}}} \; ,
\end{equation}
where $\vec{w_1}$ and $\vec{w_2}$ are obtained by flattening and concatenating the weights of the convolution filters of $C_1$ and $C_2$.

Third, we adopt the discrepancy between the two predictions $p^{(1)}$ and $p^{(2)}$ as an indicator to weight the adversarial loss. 
The self-adaptive adversarial loss can be extended from the traditional adversarial loss (Eq. \ref{eq-basic-adversarial}) as
\begin{equation}
\begin{aligned}
& \mathcal{L}_{adv} (G, D) = -E[ \log (D(G(X_S)))] \; - \\ 
& E[ (\lambda_{local}\mathcal{M}(p^{(1)}, p^{(2)})+\epsilon)\log (1-D(G(X_T))) ]  \; ,
\label{eq-sa-adversarial}
\end{aligned}
\end{equation}
where $p^{(1)}$ and $p^{(2)}$ are predictions made by $C_1$ and $C_2$, respectively, $\mathcal{M}( \cdot, \cdot )$ denotes the cosine distance, and $\lambda_{local}$ controls the adaptive weight for adversarial loss.
Note that in Eq. \ref{eq-sa-adversarial}, to stabilize the training process, we add a small number $\epsilon$ to the self-adaptive weight.

With the above loss terms, the overall loss function of our approach can be written as
\begin{equation}
\begin{aligned}
\mathcal{L}_{CLAN} (G, D) = & \mathcal{L}_{seg}(G) + \lambda_{weight} \mathcal{L}_{weight} (G) \; + \\
&\lambda_{adv} \mathcal{L}_{adv} (G, D)  \; , \label{eq:total loss}
\end{aligned}
\end{equation}
where $\lambda_{weight}$ and $\lambda_{adv}$ denote the hyper parameters that control the relative importance of the three losses. The training objective of CLAN is
\begin{equation}
\begin{aligned}
G^*, D^* = &\arg\min_G\max_D \mathcal{L}_{CLAN} (G, D).\label{eq:objective}
\end{aligned}
\end{equation}

We solve Eq.~\ref{eq:objective} by alternating between optimizing $G$ and $D$ until $\mathcal{L}_{CLAN}(G, D)$ converges.
\begin{figure*}[t]
\centering
\includegraphics[width=0.95\linewidth]{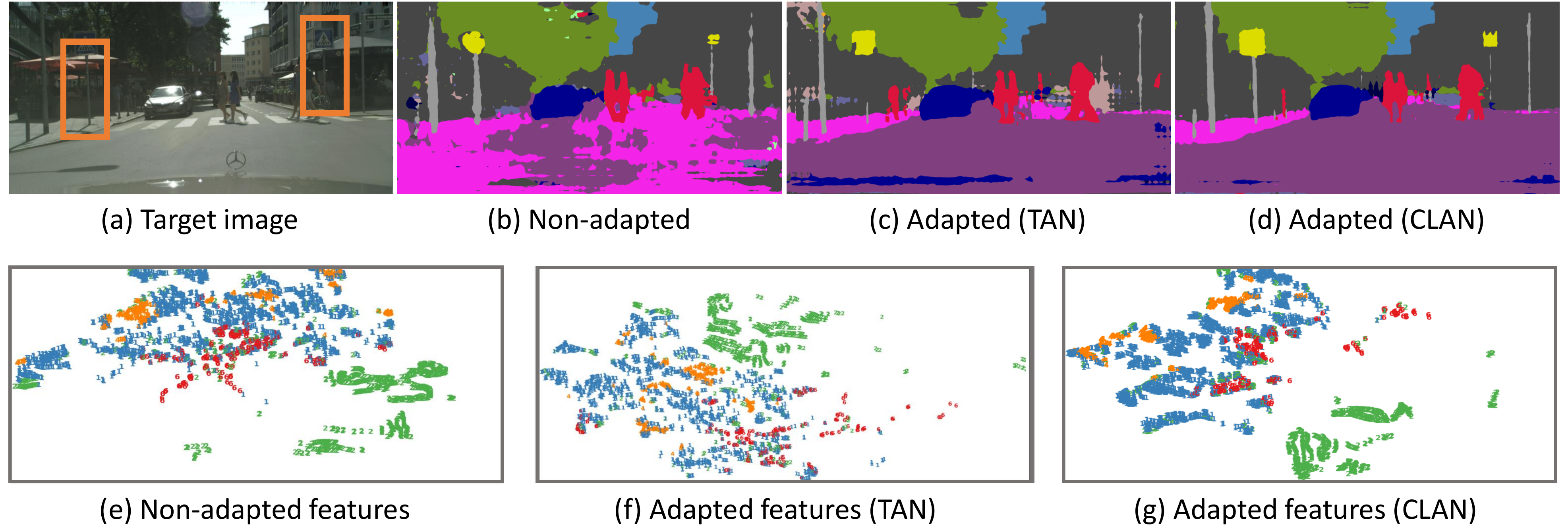}
\caption{A contrastive analysis of CLAN and traditional adversarial network (TAN). (a): A target image, and we focus on the poles and traffic signs in orange boxes. (b): A non-adapted segmentation result. Although the global segmentation result is poor, the poles and traffic signs can be correctly segmented. It indicates that some classes are originally aligned between domains, even without any domain adaptation. (c): Adapted result of TAN, in which a decent segmentation map is produced but poles and traffic signs are poorly segmented. The reason is that the global alignment strategy tends to assign a \emph{conservative prediction} to a feature and would lead some features to be predicted to other \emph{prevalent classes}~\cite{gulrajani2017improved, karras2017progressive}, thus causing those infrequent features being negatively transferred. (d): Adapted result from CLAN. CLAN reduces the weight of adversarial loss for those  aligned features. As a result, the original well-segmented class are well preserved. 
We then map the high-dimensional features of (b), (c) and (d) to a 2-D space with t-SNE~\cite{maaten2008tSNE} shown in (e), (f) and (g). The comparison of feature distributions further proves that CLAN can enforce category-level alignment during the trend of global alignment. (For a clear illustration, we only show 4 related classes, \emph{i.e.}, building in blue, traffic sign in orange, pole in red and vegetation in green.) }
\label{fig:tSNE}
\end{figure*}
\subsection{Analysis} \label{Analysis}
The major difference between the proposed framework and traditional adversarial learning consists in two aspects: the discrepancy loss and the category-level adversarial loss. Accordingly, analysis will focus on the two differences. 

First, the discrepancy (co-training) loss encourages $E$ to learn domain-invariant semantics instead of the domain specific elements such as illumination. In our network, classifiers $C_1$ and $C_2$ 1) are encouraged to capture possibly different characteristics of a feature, which is ensured by the discrepancy loss, and 2) are enforced to make the same prediction of any $E$ output (no matter the source or target), which is required by the segmentation loss and the adversarial loss. The two forces actually require that $E$ should capture the essential aspect of a pixel across the source and target domains, which, as we are aware of, is the pure semantics of a pixel, \emph{i.e.,} the domain-invariant aspect of a pixel. Without the discrepancy loss (co-training), force 1) is missing, and there is a weaker requirement for $E$ to learn domain-invariant information. On the other side, in our \emph{simulated $\rightarrow$ real} task, 
the two domains vary a lot at \emph{visual level}, but overlap at \emph{semantic level}. If $C_1$ and $C_2$ are input with \emph{visual-level features} from $E$, their predictions should be inaccurate in target domain and tend to be different, which will be \emph{punished by large adversarial losses}. As a result, once our algorithm converges, $C_1$ and $C_2$ will be input with \emph{semantic-level} features instead of \emph{visual-level} features. That is, $E$ is encouraged to learn domain-invariant semantics. Therefore, the discrepancy loss serves as an implicit contributing factor for the improved adaptation ability.

Second, in our major contribution, we extend the traditional adversarial loss with an adaptive weight $[\lambda_{local}\mathcal{M}(p^{(1)}, p^{(2)})+\epsilon]$. On the one hand, when $\mathcal{M}(p^{(1)}, p^{(2)})$ is large, feature maps of the same class do not have similar joint distributions between two domains: they suffer from the semantic inconsistency. Therefore, the weights are such assigned as to encourage $G$ to fool $D$ mainly on features that suffer from domain shift. On the other hand, when $\mathcal{M}(p^{(1)}, p^{(2)})$ is small, the joint distribution would have a large overlap across domains, indicating that the semantic inconsistency problem is not severe. Under this circumstance, $G$ tends to ignore the adversarial punishment from $D$. From the view of $D$, the introduction of the adaptive weight encourages $D$ to distill more knowledge from examples suffering from semantic inconsistency rather than those well-aligned classes. As a result, CLAN is able to improve category-level alignment degree in adversarial training. This could be regarded as an explicit contributing factor for the adaptation ability. We additionally give a contrastive analysis between traditional adversarial network (TAN) and CLAN on their adaptation result in Fig.~\ref{fig:tSNE}.
\begin{table*}[t]
\caption{
    Adaptation from GTA5~\cite{richter2016gta5} to Cityscapes~\cite{cordts2016cityscapes}. We present per-class IoU and mean IoU. ``V'' and ``R'' represent the VGG16-FCN8s and ResNet101 backbones, respectively. ``ST'' and ``AT'' represent two lines of method, \emph{i.e.,} self training- and adversarial learning-based DA. We highlight the best result in each column in \textbf{bold}. To clearly showcase the effect of CLAN on infrequent classes, we highlight these classes in \textcolor{blue}{blue}. \emph{Gain} indicates the mIoU improvement over using the source only.
    }
  \begin{center}
  \scriptsize
  \setlength{\tabcolsep}{3.3pt}
  \begin{tabular}{l|c|c|ccccccccccccccccccccc}
    \toprule
    \multicolumn{23}{c}{\textbf{GTA5 $\rightarrow$ Cityscapes}} \\
    \midrule
     &\rot{Arch.} &\rot{Meth.} & \rot{road} & \rot{side.} & \rot{buil.} & \textcolor{blue}{\rot{wall}} & \textcolor{blue}{\rot{fence}} & \textcolor{blue}{\rot{pole}} & \textcolor{blue}{\rot{light}} & \textcolor{blue}{\rot{sign}} & \rot{vege.} & \textcolor{blue}{\rot{terr.}} & \rot{sky} & \rot{pers.} & \textcolor{blue}{\rot{rider}} & \rot{car} & \textcolor{blue}{\rot{truck}} & \textcolor{blue}{\rot{bus}} & \textcolor{blue}{\rot{train}} & \textcolor{blue}{\rot{motor}} & \rot{bike} & \rot{\textbf{mIoU}} & \rot{\textbf{gain}}\\ 
     \midrule
     \midrule
     Source only & V & - & 64.0 & 22.1 & 68.6 & 13.3 & 8.7 & 19.9 & 15.5 & 5.9 & 74.9 & 13.4 & 37.0 & 37.7 & 10.3 & 48.2 & 6.1 & 1.2 & 1.8 & 10.8 & 2.9 & 24.3 & ---\\
     CBST~\cite{zou2018unsupervised} & V & ST & 90.4 & 50.8 & 72.0 & 18.3 & 9.5 & 27.2 & 28.6 & 14.1 & 82.4 & 25.1 & 70.8 & 42.6 & 14.5 & 76.9 & 5.9 & 12.5 & 1.2 & 14.0 & 28.6 & 36.1 & 11.8\\
     \midrule
     \midrule
     Source only & V & - & 25.9 & 10.9 & 50.5 & 3.3 & 12.2 & 25.4 & \bf 28.6 & 13.0 & 78.3 & 7.3 & 63.9 & \bf 52.1 & 7.9 & 66.3 & 5.2 & 7.8 & 0.9 & 13.7 & 0.7 & 24.9 & ---\\

     MCD~\cite{saito2017maximum} & V & AT & 86.4 & 8.5 & 76.1 & 18.6 & 9.7 & 14.9 & 7.8 & 0.6 & \bf 82.8 & 32.7 & 71.4 & 25.2 & 1.1 & 76.3 & 16.1 & 17.1 & 1.4 & 0.2 & 0.0 & 28.8 & 3.9\\
     \midrule
     Source only & V & - & 18.1 & 6.8 & 64.1 & 7.3 & 8.7 & 21.0 & 14.9 & \bf 16.8 & 45.9 & 2.4 & 64.4 & 41.6 & \bf 17.5 & 55.3 & 8.4 & 5.0 & \bf 6.9 & 4.3 & 13.8 & 22.3 & --- \\
     CDA~\cite{zhang2017curriculum} & V & AT & 74.9 &22.0 &71.7 &6.0 &11.9 &8.4 &16.3 &11.1 & 75.7 & 13.3 & 66.5 & 38.0 & 9.3 & 55.2 & 18.8 & 18.9 & 0.0 & \bf 16.8 & \bf 14.6 & 28.9 & 6.6\\
     \midrule
    Source only & V & - & 26.0 & 14.9 & 65.1 &  5.5 & 12.9 &  8.9 &  6.0 &  2.5 & 70.0 &  2.9 & 47.0 & 24.5 &  0.0 & 40.0 & 12.1 &  1.5 &  0.0 &  0.0 &  0.0 & 17.9 & ---\\
    FCNs in the wild~\cite{hoffman2016fcns} & V & AT & 70.4 & 32.4 & 62.1 & 14.9 &  5.4 & 10.9 & 14.2 &  2.7 & 79.2 & 21.3 & 64.6 & 44.1 &  4.2 & 70.4 &  8.0 &  7.3 &  0.0 &  3.5 &  0.0 & 27.1 & 9.2  \\
    
	CyCADA (feature)~\cite{hoffman2017cycada} & V & AT & 85.6 & 30.7 & 74.7 & 14.4 & 13.0 & 17.6 & 13.7 & 5.8 & 74.6 & 15.8 & 69.9 & 38.2 & 3.5 & 72.3 & 16.0 & 5.0 & 0.1 & 3.6 & 0.0 & 29.2 & 11.3\\ 
    
    Baseline (TAN)~\cite{tsai2018OutputSpace} & V & AT & 87.3 & 29.8 & 78.6 & 21.1 &  18.2 & 22.5 & 21.5 & 11.0 & 79.7 & 29.6 & 71.3 & 46.8 & 6.5 & 80.1 & 23.0 & 26.9 &0.0 & 10.6 & 0.3 & 35.0 & 17.1\\

    CLAN & V & AT & \bf 88.0 & \bf 30.6 & \bf 79.2 & \bf 23.4 & \bf 20.5 & \bf 26.1 & 23.0 & 14.8 & 81.6 & \bf 34.5 & \bf 72.0 & 45.8 & 7.9 & \bf 80.5 & \bf 26.6 & \bf 29.9 & 0.0 & 10.7 & 0.0 & \bf 36.6 & \bf 18.7\\
		\midrule
        \midrule
	Source only & R & - & 75.8 & 16.8 & 77.2 & 12.5 & 21.0 & 25.5 & 30.1 & 20.1 & 81.3 & 24.6 & 70.3 & 53.8 & 26.4 & 49.9 & 17.2 & 25.9 & 6.5 & 25.3 & 36.0 & 36.6 & ---\\
    
	Baseline (TAN)~\cite{tsai2018OutputSpace} & R & AT &  86.5 & 25.9 & \bf 79.8 & 22.1 & 20.0 & 23.6 & 33.1 &  21.8 & 81.8 & 25.9 & \bf 75.9 & 57.3 & 26.2 & \bf 76.3 & 29.8 & 32.1 & \bf  7.2 & 29.5 & \bf 32.5 & 41.4 & 4.8\\
    
	CLAN & R & AT & \bf 87.0 & \bf 27.1 & 79.6 & \bf27.3 & \bf 23.3 &	\bf 28.3 & \bf 35.5 & \bf 24.2 & \bf 83.6 & \bf 27.4 & 74.2 & \bf 58.6 & \bf 28.0 & 76.2 &   \bf 33.1 & \bf 36.7 & 6.7 & \bf 31.9 & 31.4 & \bf 43.2 &	\bf 6.6\\ 
    \bottomrule
  \end{tabular}
  \end{center}
  \label{table:gta-cityscapes}
\end{table*}

\begin{table*}[t]
  \caption{
    Adaptation from SYNTHIA~\cite{ros2016synthia} to Cityscapes~\cite{cordts2016cityscapes}. We present per-class IoU and mean IoU for evaluation. CLAN and state-of-the-art domain adaptation methods are compared. For each backbone, the best accuracy is highlighted in \textbf{bold}. To clearly showcase the effect of CLAN on infrequent classes, we highlight these classes in \textcolor{blue}{blue}. \emph{Gain} indicates the mIoU improvement over using the source only.
    }
  \begin{center}
  \scriptsize
  \setlength{\tabcolsep}{6.45pt}
  \begin{tabular}{l|c|c|ccccccccccccccc}
    \toprule
    \multicolumn{17}{c}{\textbf{SYNTHIA $\rightarrow$ Cityscapes}} \\
    \midrule
     &\rot{Arch.} & \rot{Meth.} & \rot{road} & \rot{side.} & \rot{buil.} & \textcolor{blue}{\rot{light}} & \textcolor{blue}{\rot{sign}} & \rot{vege.} & \rot{sky} & \rot{pers.} & \textcolor{blue}{\rot{rider}} & \rot{car} & \textcolor{blue}{\rot{bus}} & \textcolor{blue}{\rot{motor}} & \rot{bike} & \rot{\textbf{mIoU}} & \rot{\textbf{gain}}
     \\ 
     \midrule
     \midrule
    Source only & V & - & 17.2 & 19.7 & 47.3 & 3.0 & 9.1 & 71.8 & 78.3 & 37.6 & 4.7 & 42.2 & 9.0 & 0.1 & 0.9 & 26.2 & --- \\
    
    CBST~\cite{zou2018unsupervised} & V & ST & 69.6 & 28.7 & 69.5 & 11.9 & 13.6 & 82.0 & 81.9 & 49.1 & 14.5 & 66.0 & 6.6 & 3.7 & 32.4 & 36.1 & 9.9\\
    
    \midrule
    \midrule
    
    Source only & V & - & 6.4 & 17.7 & 29.7 & 0.0 & 7.2 & 30.3 &  66.8 &  51.1 & 1.5 &  47.3 & 3.9 & 0.1 & 0.0 & 20.2 & ---\\
    
    FCNs in the wild~\cite{hoffman2016fcns} & V & AT & 11.5 & 19.6 & 30.8 & 0.1 & \bf 11.7 & 42.3 & 68.7 & 51.2 & 3.8 & 54.0 & 3.2 & 0.2 & 0.6 & 22.9 & 2.7 \\

	Cross-city~\cite{chen2017cross} & V & AT & 62.7 & 25.6 & \bf 78.3 & 1.2 & 5.4 & \bf 81.3 & \bf 81.0 & 37.4 & 6.4 & 63.5 & 16.1 & 1.2 & 4.6 & 35.7 & 15.2\\ 

    Baseline (TAN)~\cite{tsai2018OutputSpace} & V & AT & 78.9 & 29.2 & 75.5 & 0.1 & 4.8 & 72.6 & 76.7 & 43.4 & 8.8 & 71.1 & 16.0 & \bf 3.6 & 8.4 & 37.6 & 17.4\\
    
    CLAN & V & AT & \bf 80.4 & \bf 30.7 & 74.7 & \bf 1.4 & 8.0 & 77.1 & 79.0 & \bf 46.5 & \bf 8.9 & \bf 73.8 & \bf 18.2 & 2.2 & \bf 9.9 & \bf 39.3 & \bf 19.1\\
	
	\midrule
	\midrule
        
	Source only & R & - & 55.6 & 23.8 & 74.6 & 6.1 & 12.1 & 74.8 & 79.0 & 55.3 & 19.1 & 39.6 & 23.3 & 13.7 & 25.0 & 38.6 & ---\\
    
	Baseline (TAN)~\cite{tsai2018OutputSpace} & R & AT & 79.2 & \bf 37.2 & 78.8 & 9.9 & 10.5 & \bf 78.2 & 80.5 & \bf 53.5 & 19.6 & 67.0 & 29.5 & 21.6 & \bf 31.3 & 45.9 & 7.3 \\
    
	CLAN & R & AT & \bf 81.3 & 37.0 & \bf 80.1 & \bf 16.1 &	\bf 13.7 &	\bf 78.2 & \bf 81.5 & 53.4 & \bf 21.2 & \bf 73.0 & \bf 32.9 & \bf 22.6 & 30.7 & \bf 47.8 & \bf 9.2\\ 
    \bottomrule
  \end{tabular}
  \end{center}
  \label{table:synthia-cityscapes}
  \vspace{-3mm}
\end{table*}

\section{Experiment}
\subsection{Datasets}
We evaluate CLAN together with several state-of-the-art algorithms on two adaptation tasks, \emph{e.g.}, SYNTHIA~\cite{ros2016synthia} $\rightarrow$ Cityscapes~\cite{cordts2016cityscapes} and GTA5~\cite{richter2016gta5} $\rightarrow$ Cityscapes. Cityscapes is a real-world dataset with 5,000 street scenes. We use Cityscapes as the target domain. GTA5 contains 24,966 high-resolution images compatible with the Cityscapes annotated classes. SYNTHIA contains 9400 synthetic images. We use SYNTHIA or GTA5 as the source domain. 

\subsection{Implementation Details}
We use PyTorch for implementation. We utilize the DeepLab-v2~\cite{chen2018deeplab} framework with ResNet-101~\cite{he2016resnet} pre-trained on ImageNet~\cite{deng2009imagenet} as our source-only backbone for network $G$. We use the single layer adversarial DA method proposed in~\cite{tsai2018OutputSpace} as the TAN baseline. For co-training, we duplicate two copies of the last classification module and arrange them in parallel after the feature extractor, as illustrated in Fig.~\ref{fig:main}. For a fair comparison to those methods with the VGG backbone, we also apply CLAN on VGG-16 based FCN8s~\cite{long2015fcn}. For network $D$, we adopt a similar structure with~\cite{radford2015unsupervised}, which consists of 5 convolution layers with kernel $4 \times 4$ with channel numbers $\{64, 128, 256, 512, 1\}$ and stride of $2$. Each convolution layer is followed by a Leaky-ReLU~\cite{maas2013ReLU} parameterized by $0.2$ except the last layer. Finally, we add an up-sampling layer to the last layer to rescale the output to the size of the input map, in order to match the size of local alignment score map. During training, we use SGD~\cite{bottou2010SGD} as the optimizer for $G$ with a momentum of $0.9$, while using Adam~\cite{kingma2014adam} to optimize $D$ with $\beta_1 = 0.9$, $\beta_2 = 0.99$. We set both optimizers a weight decay of $5e-4$. For SGD, the initial learning rate is set to $2.5e-4$ and decayed by a poly learning rate policy, where the initial learning rate is multiplied by $(1 - \frac{iter}{max\_iter})^{power}$ with $power = 0.9$. For Adam, we initialize the learning rate to $5e-5$ and fix it during the training. We train the network for a total of $100k$ iterations. We use a crop of $512 \times 1024$ during training, and for evaluation we up-sample the prediction map by a factor of 2 and then evaluate mIoU. In our best model, the hyper-parameters $\lambda_{weight}$, $\lambda_{adv}$, $\lambda_{local}$ and $\epsilon$ are set to $0.01$, $0.001$, $40$ and 0.4 respectively.

\subsection{Comparative Studies}
\begin{figure*}[ht]
\centering
\includegraphics[width=0.95\linewidth]{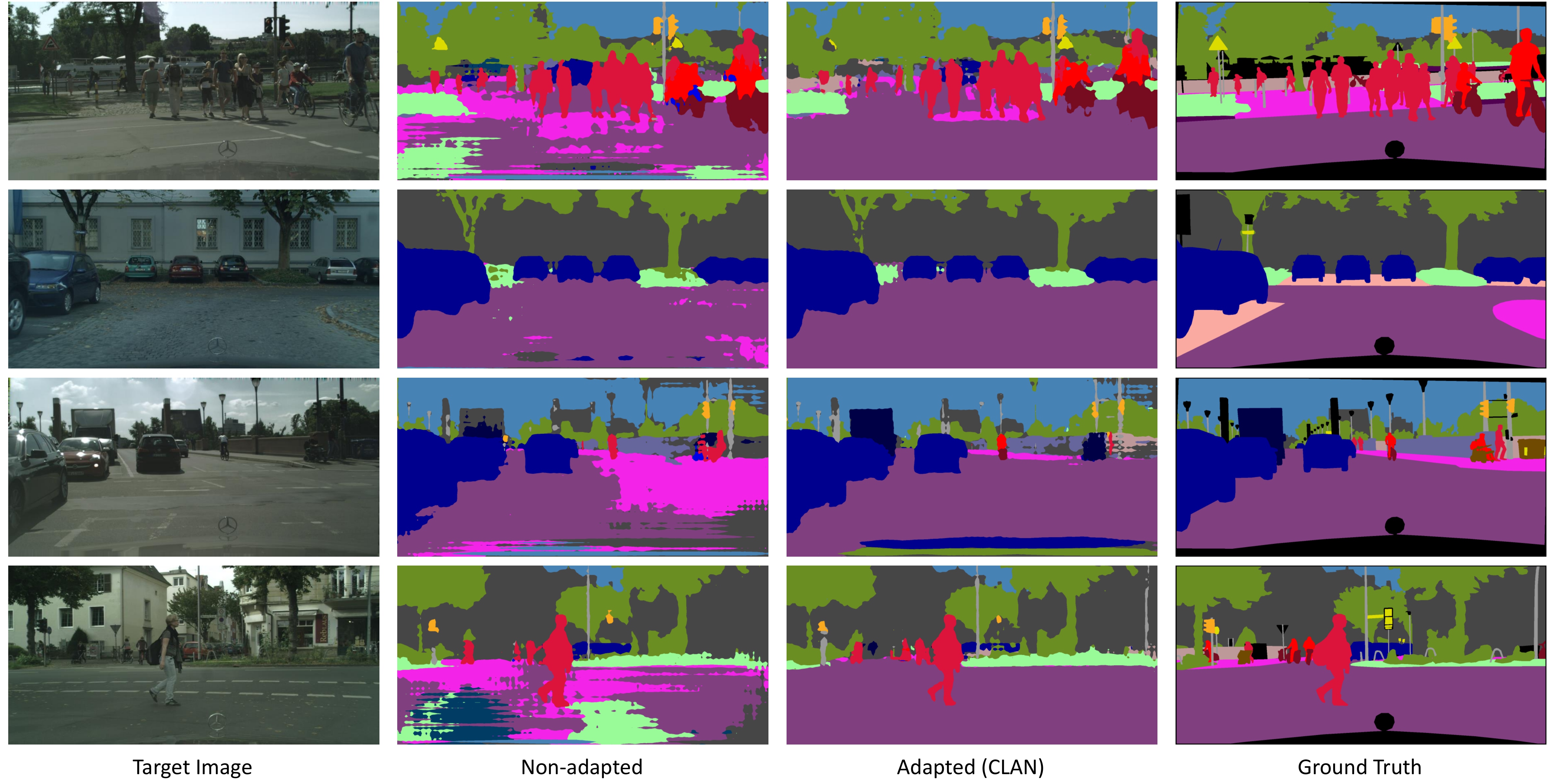}
\caption{Qualitative results of UDA segmentation for GTA5 $\rightarrow$ Cityscapes. For each target image, we show the non-adapted (source only) result, adapted result with CLAN and the ground truth label map, respectively.}
\label{fig:result}
\end{figure*}

We present the adaptation results on task GTA5 $\rightarrow$ Cityscapes in Table~\ref{table:gta-cityscapes} with comparisons to the state-of-the-art domain adaptation methods~\cite{saito2017maximum,zhang2017curriculum,hoffman2016fcns,hoffman2017cycada,tsai2018OutputSpace,zou2018unsupervised}. We observe that CLAN significantly outperforms the source-only segmentation method by $+18.7\%$ on VGG-16 and $+6.6\%$ on ResNet-101. Besides, CLAN also outperforms the state-of-the-art methods, which improves the mIOU by over $+7\%$ compared with MCD~\cite{saito2017maximum}, CDA~\cite{zhang2017curriculum} and CyCADA~\cite{hoffman2017cycada}. Compared to traditional adversarial network (TAN) in the output space~\cite{tsai2018OutputSpace}, CLAN brings over $+1.6\%$ improvement in mIOU in both architectures of VGG-16 and ResNet-101. In some infrequent classes which are prone to suffer from the side effect of global alignment, \emph{e.g.,} fence, traffic light and pole, CLAN can significantly outperform TAN. Besides, we also compare CLAN with the self training-based methods, among which CBST~\cite{zou2018unsupervised} is the current state-of-the-art one. This series of explicit methods usually achieve higher mIoU then the implicit feature alignment. While in our experiment, we find that CLAN is on par with CBST. Some qualitative segmentation examples can be viewed in Fig. \ref{fig:result}.

Table~\ref{table:synthia-cityscapes} provides the comparative results on the task SYNTHIA $\rightarrow$ Cityscapes. On VGG-16, our final model yields $39.3\%$ in terms of mIOU, which significantly improves the non-adaptive segmentation result by $19.1\%$. Besides, CLAN outperforms the current state-of-art method~\cite{hoffman2016fcns} by $16.4\%$ and~\cite{chen2017cross} by $3.6\%$. On ResNet-101, CLAN brings $9.2\%$ improvement to source only segmentation model. Compare to TAN~\cite{tsai2018OutputSpace}, the use of adaptive adversarial loss also brings $1.9\%$ gain in terms of mIOU. Likewise, CLAN is more effective on those infrequent classes which are prone to be negatively transferred, such as traffic light and sign, bringing over $3.2\%$ improvement respectively. While on some prevalent classes, CLAN can also be par on with the baseline method. Note that on the ``train'' class, the improvement is not stable. This is due to the training samples that contain the ``train'' are very few. Finally, comparing with the self training-based method, CLAN outperforms CBST by $3.2\%$ in terms of mIOU. These observations are in consistent with our t-SNE analysis in Fig.~\ref{fig:tSNE}, which further verifies that CLAN can actually boost the category-level alignment in segmentation-based DA task.

\begin{figure}
\begin{minipage}[b]{.49\linewidth}
\centering
\includegraphics[height=3.0cm]{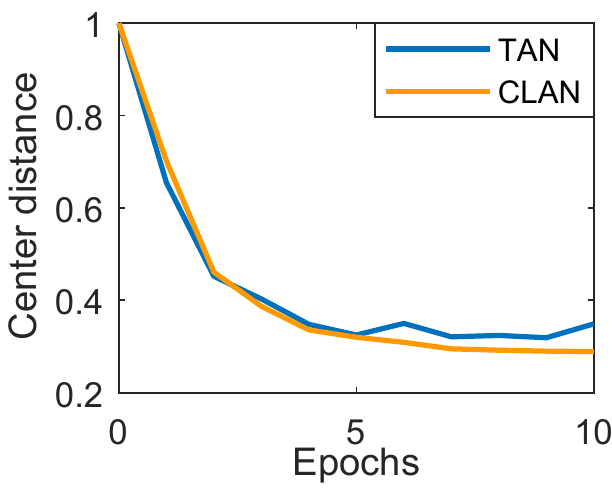}
\end{minipage}
\begin{minipage}[b]{.49\linewidth}
\centering
\includegraphics[height=3.0cm]{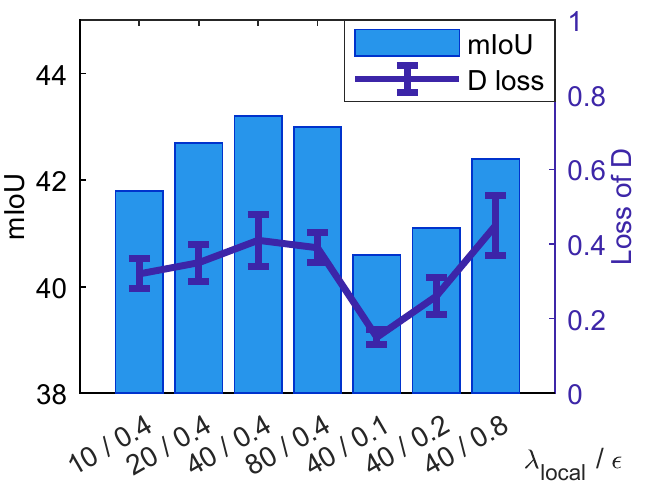}
\end{minipage}
\caption{\textbf{Left:} Cluster center distance variation as training goes on. \textbf{Right:} Mean IoU (see bars \& left y axis) and convergence performance (see lines \& right y axis) variation when training with different $\lambda_{local}$ and $\epsilon$.}
\label{fig:two}
\end{figure}

\subsection{Feature Distribution}
To further verify that CLAN is able to decrease the negative transfer effect for those well-aligned features, we designed an experiment to take a closer look at the category-level alignment degree of each class. Specifically, we randomly select 1K source and 1K target images and calculate the cluster center distance (CCD) $\{d_1^e ... d_n^e\}$ of features of the same class between two domains, where $n = \#class$ and $e$ is training epoch. $d_i^e$ is normalized by $d_i^e / d_i^0$ (In this way, the CCD from the pre-trained model without any fine-tuning would be always normalized to 1). We report $d_i^e$ in Fig.~\ref{fig:two} (left subfigure, taking the class ``wall'' as an example). First, we observe as training goes on, $d_i^e$ is monotonically decreasing in CLAN while not being monotone in TAN, suggesting CLAN prevents the well-aligned features from being incorrectly mapped. Second, $d_i^e$ converges to a smaller value in CLAN than TAN, suggesting CLAN achieves better feature alignment at semantic level. 

\begin{figure*}[ht]
\centering
\includegraphics[width=0.95\linewidth]{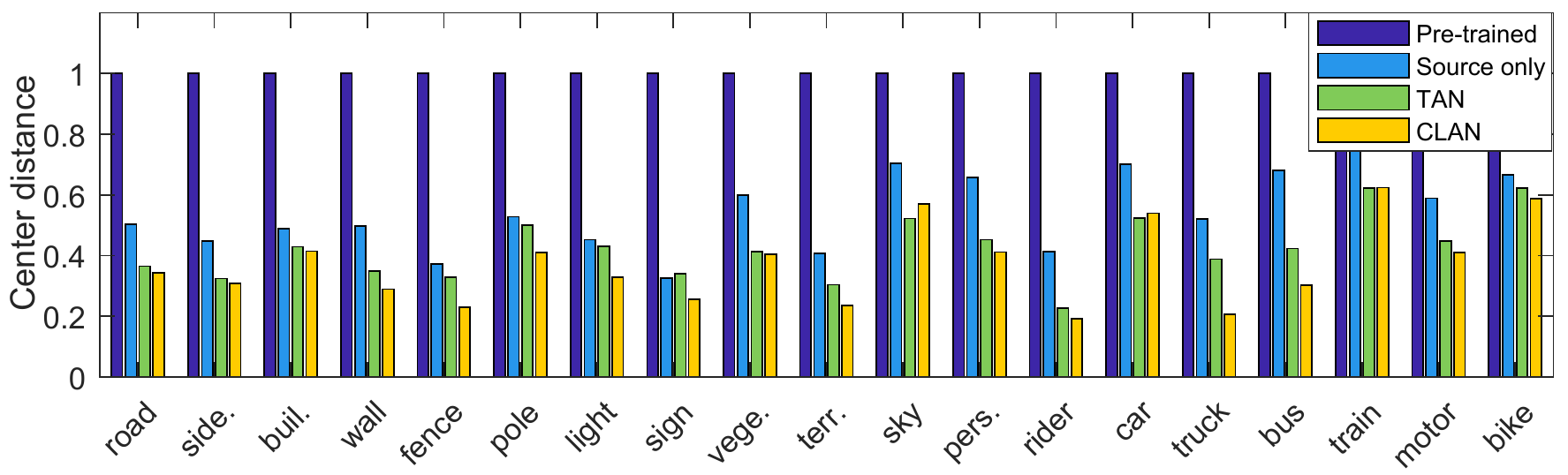}
\caption{Quantitative analysis of the feature joint distributions. For each class, we show the distance of the feature cluster centers between source domain and target domain. These results are from 1) the model pre-trained on ImageNet~\cite{deng2009imagenet} without any fine-tuning, 2) the model fine-tuned with source images only, 3) the adapted model using TAN and 4) the adapted model using CLAN, respectively.}
\label{fig:bar}
\end{figure*}

We further report the final CCD of each class in Fig.~\ref{fig:bar}. We can observe that CLAN can achieve a smaller CCD in most cases, especially in those infrequent classes which are prone to be negatively transferred. These quantitative results, together with the qualitative t-SNE~\cite{maaten2008tSNE} analysis in Fig.~\ref{fig:tSNE}, indicate that CLAN can preferably align the two domains in semantic level. Such category-aligned feature distribution usually makes the subsequent classification easier.

\subsection{Parameter Studies}\label{sec:parameter}
In this experiment, we aim to study two problems: 1) whether the adaptive adversarial loss would cause instability (vanishing gradient) during adversarial training and 2) how much the adaptive adversarial loss would effect the performance. For the problem 1), we utilize the loss of $D$ to indicate the convergence performance and a stable adversarial training is achieved if $D$ loss converges around 0.5. First, we test our model using $\lambda_{local} = 40$, with varying $\epsilon$ over a range \{0.1, 0.2, 0.4, 0.8\}. We do not use any $\epsilon$ larger than 0.8 since CLAN would degrade into TAN in that case.
In the experiment, our model suffers from poor convergence when utilize a very small $\epsilon$, \emph{e.g.}, 0.1 or 0.2. It indicates that a proper choice of $\epsilon$ is between 0.2 and 0.8. Motivated by this observation, we then test our model using $\epsilon = 0.4$ with varying $\lambda_{local}$ over a range \{10, 20, 40, 80\}. We observe that the convergence performance is not very sensitive to $\lambda_{local}$ since the loss of $D$ converges to proper values in all the cases. The best performance is achieved when using $\lambda_{local} = 40$ and $\epsilon = 0.4$. Besides, we observe that the adaptation performance of CLAN can steadily outperform TAN when using parameters near the best value. We present the detailed performance variation in Fig.~\ref{fig:two} (right subfigure). By comparing both the convergence and segmentation results under these different parameter settings, we can conclude that our proposed adaptive adversarial weight can significantly effect and improve the adaptation performance.


\section{Conclusion}
In this paper, we introduce the category-level adversarial network (CLAN), aiming to address the problem of semantic inconsistency incurred by global feature alignment during unsupervised domain adaptation (UDA). By taking a close look at the category-level data distribution, CLAN adaptively weight the adversarial loss for each feature according to how well their category-level alignment is. In this spirit, each class is aligned with an adaptive adversarial loss. Our method effectively prevents the well-aligned features from being incorrectly mapped by the side effect of pure global distribution alignment. Experimental results validate the effectiveness of CLAN, which yields very competitive segmentation accuracy compared with state-of-the-art UDA approaches.

\textbf{Acknowledgment.} This work is partially supported by the National Natural Science Foundation of China (No. 61572211).

{\small
\bibliographystyle{ieee}
\bibliography{egpaper_final}
}

\end{document}